\pdfoutput=1

\documentclass[11pt]{article}
\usepackage[]{acl}

\usepackage{times}
\usepackage{dsfont}
\usepackage{latexsym}
\usepackage{url}
\usepackage[T1]{fontenc}
\usepackage{graphicx}
\usepackage{multirow}
\usepackage{amsmath}
\usepackage{amssymb}
\usepackage{bbm}
\usepackage{pifont}
\usepackage{adjustbox}
\usepackage{tabularx}
\usepackage{tablefootnote}
\usepackage[utf8]{inputenc}
\usepackage{subfigure}
\usepackage{parskip}
\usepackage{caption}
\usepackage{enumitem}
\usepackage{color}
\usepackage{comment}
\usepackage{booktabs}
\usepackage{microtype}
\usepackage{makecell}
\usepackage{bbding}
\usepackage{algorithm}
\usepackage{algpseudocode}
\usepackage[T1]{fontenc}

\usepackage[utf8]{inputenc}

\usepackage{microtype}

\title{Early Rumor Detection Using Neural Hawkes Process with a New Benchmark Dataset}

\author{Fengzhu Zeng \\
   	Singapore Management University \\
   	81 Victoria St, Singapore 188065 \\
   \texttt{fzzeng.2020@phdcs.smu.edu.sg} \And
   Wei Gao\\
   Singapore Management University \\
   81 Victoria St, Singapore 188065 \\
   \texttt{weigao@smu.edu.sg}}

\begin{document}
\maketitle 
\begin{abstract}
Little attention has been paid on \underline{EA}rly \underline{R}umor \underline{D}etection (EARD), and EARD performance was evaluated inappropriately on a few datasets where the actual early-stage information is largely missing. To reverse such situation, we construct BEARD, a new \underline{B}enchmark dataset for \underline{EARD}, based on claims from fact-checking websites by trying to gather as many early relevant posts as possible.
We also propose HEARD, a novel model based on neural \underline{H}awkes process for \underline{EARD}, which can guide a generic rumor detection model to make timely, accurate and stable predictions. Experiments show that HEARD achieves effective EARD performance on two commonly used general rumor detection datasets and our BEARD dataset.
\end{abstract}

\section{Introduction}

The proliferation of online rumors has aroused widespread concerns. Many studies have been conducted for automatic rumor detection in social media and achieved high detection accuracy~\citep{ma2016detecting,ma2017detect,yu2017convolutional,ruchansky2017csi,ma-etal-2018-rumor,guo2018rumor,bian2020rumor}. However, these generic detection methods lack in-depth modeling of temporality,
which can cause tremendous delay in detection given the instantaneous and sporadic nature of rumor propagation.

\begin{table}[t!]
\centering
\normalsize
\begin{adjustbox}{width={\linewidth},totalheight={\textheight},keepaspectratio}%
\begin{tabular}{lll}
\toprule                                                 
 &\textbf{Time} & \textbf{Post}                                                      \\ 
\midrule
\multirow{2}{*}{T.ORG.} &\textit{2015-07-07}  & \textit{Translucent butterfly - beautiful!}         \\ 
& {2015-10-01}          & \#Snopes  Translucent Butterfly URL\\ 
\midrule
\multirow{4}{*}{T.REC.}&{\textit{2013-02-08}} & \textit{Ever see a translucent butterfly?}                                 \\
&{2013-06-24}           & fake...like bubbles              \\
&{2014-09-03}          & ...fake, a Worth1000 Photoshop contest entry        \\
&{2014-10-06}          & Multiple repeat fakes                                   \\
\midrule
\midrule
\multirow{2}{*}{P.ORG.} &\textit{01-08 00:00}  & \textit{… courtesy of Banksy.}         \\ 
& {01-08 14:39}          & ...it is not by Banksy its by @LucilleClerc\\ 
\midrule
\multirow{5}{*}{P.REC.}&{\textit{01-07 20:09}} & \textit{Banksy for Charlie...}                                 \\
&{01-07 23:48}           &  That’s not Banksy though, just someone fan page              \\
&{01-08 00:14}         & …I don't think that Banksy Insta is official is it?                                   \\
&{01-08 08:33}          & The person (not B.) shared it from another source.         \\
&{01-08 09:55}          & Not Banksy btw. It's @LucilleClerc                             \\ 
\bottomrule
\end{tabular}
\end{adjustbox}
\caption{A rumor in TWITTER dataset claiming "A butterfly with translucent wings" and another from PHEME claiming "A street artist Banksy posted an illustration on the Instagram as tribute to Charlie Hebdo". T. (P.) denotes TWITTER (PHEME) dataset, and ORG. (REC.) denotes the original (recollected) set of posts. The earliest post in each set is in \textit{Italic}.}
\label{tab:intro}
\end{table}

While a few EArly Rumor Detection (EARD) models have been proposed~\citep{LiuW18early,zhou-etal-2019-early,song2019ced,xia-etal-2020-state}, they have been designed with oversimplification and evaluated inappropriately using the datasets constructed for generic rumor detection. Widely used rumor detection datasets, such as TWITTER~\cite{ma2016detecting} and PHEME~\cite{zubiaga2016learning}, are generally limited in covering relevant posts in the early stage as there was no mechanism ensuring to gather information that is further away from the official debunking time of a rumor. 
For this reason, the generalizability of EARD cannot be effectively trained nor be genuinely reflected using a general rumor detection dataset. 
As an example, we showcase two rumors from TWITTER and PHEME datasets in Table~\ref{tab:intro}. We manually trace Twitter conversations about each claim and recollect as many early posts relevant to it as we could. It is observed that the original posts in both datasets are clearly delayed as compared to our recollected posts. Also, the rumor indicative patterns in the recollected posts unfold differently, where dissenting voices, a common indicator of rumor, appear much earlier and may last for many hours or even years, evolving from vaguely opposing the claim (e.g., `like bubbles', `someone') to firmly refuting it with evidence (e.g., `Photoshop contest'). The original ``early'' posts in PHEME clearly fail to cover such useful patterns reflecting the early dynamics, and the posts in TWITTER do not cover any early indicative signals before the rumor was officially debunked by Snopes. 
Given the unavailability of EARD-specific dataset, it is necessary to construct a \underline{B}enchmark dataset for \underline{EARD} (BEARD) considering the earliness of relevant posts to gather.

Meanwhile, EARD methods have not been well studied either in the literature. Prior works claimed as being able to do early detection can be divided into two categories, both of which are sub-optimal: 1) Methods that are unable to automatically determine a time point for confirming the detection~\cite{zhao2015enquiring,nguyen2017early,wu2017gleaning,LiuW18early,xia-etal-2020-state}. Typically, such methods apply a generic rumor detection model to report a decision of classification (e.g., rumor or non-rumor) at each of the \emph{pre-determined} checkpoints while leaving the determination of the best detection point to human judge. This is subject to a delayed decision as the results at the later checkpoints also need to be examined.
2) Methods that are trained to automatically determine an early detection point, but cannot guarantee the stability of decision~\cite{zhou-etal-2019-early,song2019ced}. For example, CED~\cite{song2019ced} decides an early detection point using a fixed probability threshold to assess if the current prediction is credible or not. However, prediction probability does not really reflect model's confidence~\cite{guo2017calibration}, and such decision without properly modeling the \emph{uncertainty beyond the decision point} may fail to give a timely and reliable detection because the prediction could flip over and over again afterwards with new posts flow in. 

In this work, we propose a new method called Hawkes EArly Rumor Detection (HEARD) to model the stabilization process of rumor detection based on a Neural Hawkes Process (NHP)~\cite{mei-etal-2017-The}, which can automatically determine when to make a timely and stable decision of detection. The basic idea is to construct a detection stability distribution over the \emph{expected future} predictions based on a sequence of \emph{prior and current} predictions, such that an optimal time point can be fixed without any delay for awaiting and checking the upcoming data beyond that point. 
Our main contributions can be summarized as follows\footnote{Dataset and source code are released at \url{https://github.com/znhy1024/HEARD}}:
\begin{itemize}[leftmargin=*]
    \item We introduce BEARD, the first EARD-oriented dataset, collected by covering as much as possible the early-stage information relevant to the concerned claims.
    \item We propose HEARD, a novel EARD model based on the NHP to automatically determine an optimal time point for the stable decision of early detection. 
    \item Extensive experiments show that HEARD achieves more effective EARD performance as compared to strong baselines on BEARD and two commonly used general rumor detection datasets.
\end{itemize}

\section{Related Work}
\subsection{Early Rumor Detection}
Despite extensive research on general rumor detection, early detection has not been studied well. Many studies claimed that their general detection models can be applied to early detection by simply fed with data observed up to a set of pre-determined checkpoints~\cite{ma2016detecting,yu2017convolutional,ma2017detect,ma-etal-2018-rumor,guo2018rumor,bian2020rumor}. Nevertheless, how to determine an optimal early detection point from many checkpoints is missing and non-trivial, as deciding when to stop often needs to check the data or model's outputs after the current checkpoint, causing delays of detection. 

Some methods were claimed further as designed for early detection. \citet{zhao2015enquiring} proposed to gather related posts with skeptical phrases, and performed detection with cluster-based classifiers over real-time posts. \citet{nguyen2017early} developed a hybrid neural model for post-level representation and credit classification, which were incorporated with the temporal variations of handcrafted features for detecting rumors. \citet{wu2017gleaning} clustered relevant posts and selected key features from clusters to train a topic-independent classifier for revealing emergent rumors. 
\citet{xia-etal-2020-state} employed burst detection to segment an event into sub-events and trained an encoder for each sub-event representation for incremental prediction. None of the above methods really address the key issues of early detection as they lack mechanisms enforcing the earliness, and they cannot automatically fix an optimal detection point either. 

ERD~\cite{zhou-etal-2019-early} used deep reinforcement learning to enforce model to focus on early time intervals for the trade-off between accuracy and earliness of detection, and is the first EARD method that can automatically decide to stop or continue at a checkpoint. \citet{song2019ced} proposed another EARD method called Credible Detection Point (CED) using a fixed probability threshold to determine if detection process should stop depending on the credibility of current prediction. However, these models are unstable or of low confidence because the uncertainty of future predictions is not taken into account in training.

\subsection{Rumor Detection Datasets}
Quite a few rumor detection datasets based on social media posts relevant to a set of claims were released, such as TWITTER~\cite{ma2016detecting}, PHEME~\cite{zubiaga2016learning}, RumourEval-2017/19~\cite{derczynski-etal-2017-semeval,gorrell-etal-2019-semeval},
FakeNewsNet~\cite{shu2020fakenewsnet}, etc..
RumourEval-2017/19 are minor variants of PHEME while FakeNewsNet is never used for EARD. These datasets were built for general detection of rumors without much consideration on the earliness of information. Thus, the actual early-stage social engagements may not be covered by different data collection mechanisms used, such as applying a rigid time cut-off in Search API or launching a real-time gathering with Streaming API \emph{after} a news outbreak. To our best knowledge, there is no dataset specifically built for early rumor detection task. 

\section{BEARD Corpus Construction} \label{corpus}
We scrape the text of title, claim, debunking time and veracity label in the articles on the fact-checking website \url{snopes.com}. Our goals are two-fold: 1) The collected posts are not only relevant to the claim but can diversely cover copious variations of relevant text expressions; 2) The collection can cover posts of early arrival, possibly ahead of the pertinent news exposure on the mainstream media. 

To this end, we firstly construct high-quality search queries for Twitter search. An original query is formed from the title and claim of each article, with stop words removed. Since the lengthy query might harm the diversity of search results, we utilize some heuristics to obtain a substantial set of variants of each query potentially with better result coverage in Twitter search: i) We preform synonym replacement to create a set of variants of the query; ii) We shorten each variant by removing its words one by one with carefully crafted rules to maintain useful information, e.g., named entities, for good search quality, while keeping the remaining words after each removal as a new variant. As a result, we obtain a substantial set of variants of the original query and merge the Twitter search results of each query and all its variants.

To cover early posts, each Twitter search is performed in an iterative fashion. To avoid ground-truth leakage, we first obtain the possible earliest official debunking time of the given claim by cross-checking its similar claims in a range of fact-checking websites (see Appendix~\ref{Claims Information Retrieval}). From the earliest debunking time, we search backward for the relevant posts within $M$ days \emph{prior to} debunking, and then push back further $N$ days earlier than before in each iteration until the number of newly gathered posts in an iteration becomes less than 1\% of the posts obtained from the previous iteration.

Finally, for each retrieved post, we use its conversation ID to find the root post of the conversation it is engaged in. We utilize Sentence-BERT~\cite{reimers-gurevych-2019-sentence} to retain those root posts with cosine similarities to the claim being higher than an empirical threshold. Thus far, we have obtained a set of conversation IDs for each claim which are led by different root posts (see Appendix~\ref{Posts Collection} for post-processing).
Then we fetch from Twitter all the posts in the detected conversations along with the root posts into our final collection as an instance, and label the conversation as rumor if the corresponding claim is from the ``Fact Checks'' category on the Snopes or non-rumor if it is from the ``News'' category.

Due to space limit, we provide the details of search queries construction in Appendix~\ref{Query Construction}, and the settings of iterative Twitter search in Appendix~\ref{Iterative Twitter search}.

\begin{figure*}[t!]
  \centering
  \includegraphics[width=\textwidth]{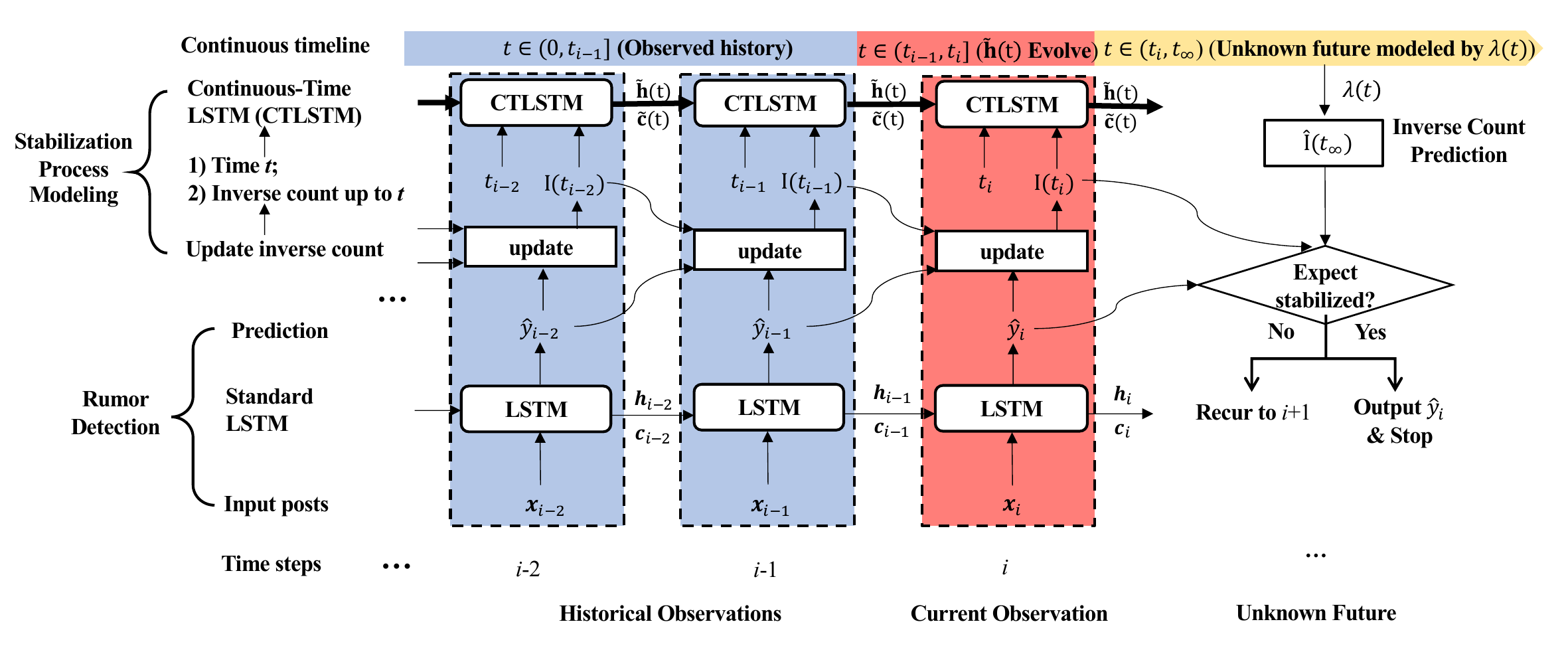}
  \caption{The architecture of HEARD. When the current observation arrives at time $t_i$, rumor detection predictions from LSTM $\hat{y}_i$ and $\hat{y}_{i-1}$ are used to update PI count $I(t_i)$. Then the intensity function $\lambda(t)$ is computed by CTLSTM for $t>t_i$. HEARD will determine $t_i$ as the earliest time point with stable prediction $\hat{y}_i$ and stop if the stability prediction $\hat{I}(t_{\infty})$ equals to $I(t_i)$, or continue otherwise.}
  \label{fig:HEARD}
\end{figure*}

\section{Problem Definition}
Let $\mathbf{C}=\left \{ C \right \}$ denote a set of instances, where each $C=\left \{y,S\right \}$ consists of the ground-truth label $y\in\{0,1\}$ and a set of relevant posts in chronological order $S=\left \{(\mathbf{m}_{1},\tau_{1}),...,(\mathbf{m}_{i},\tau_{i}),...,(\mathbf{m}_{|S|},\tau_{|S|}) \right \}$. 
$y$ indicates $C$ is a rumor if $y=1$ or a non-rumor otherwise.
$|S|$ is the number of relevant posts in $S$. 
Each tuple $(\mathbf{m}_i, \tau_i)\in S$ includes the text content $\mathbf{m}_{i}$ and the timestamp $\tau_{i}$ of the $i$-th post, where $\tau_{i}$ is defined as the time difference between the first and the $i$-th post, such that $\tau_{1}=0$ and $\tau_{i-1}\le \tau_{i}$ for $i>1$. In other words, $\tau_{i}$ can be regarded as the elapsed time relative to the earliest post so that the timelines of different instances are aligned.

Following the pre-processing method in most prior studies~\cite{ma2016detecting,song2019ced,zhou-etal-2019-early}, we divide each posts sequence into a sequence of intervals to avoid excessively long sequence. We chop a sequence $S$ into intervals based on three strategies: 1) fixed posts number, 2) fixed time length and 3) variable length in each interval~\cite{zhou-etal-2019-early}. Hence $S$ is converted to $X= \{(\mathbf{x}_i,t_i)\}_{i=1}^{|X|}$, where $|X|$ is the number of intervals, $\mathbf{x}_i=\{\mathbf{m}_{i,1},\mathbf{m}_{i,2},...,\mathbf{m}_{i,|\mathbf{x}_i|}\}$ and $t_{i}=\tau_{i,|\mathbf{x}_i|}$ which is the timestamp of the \emph{last} post $\mathbf{m}_{i,|\mathbf{x}_i|}$ in the $i$-th interval. Then, we merge the posts falling into the same interval as a single post.

We define the EARD task as automatically determining the \emph{earliest} time $\hat{t}\in\{t_i\}$, such that the prediction $\hat{y}\in\{0,1\}$ at $\hat{t}$ for a given claim is \emph{accurate and remains unchanged} afterwards with time goes by. It is worthwhile to mention that since $t_i$ relates to the granularity of intervals, it might affect the precision of a decision point based on the formed intervals. In practice, however, we will try to make the intervals small for keeping such impact marginal.

\section{HEARD Model} \label{post merge}

Figure~\ref{fig:HEARD} shows the architecture of HEARD, which contains two components: 1) the rumor detection component predicts rumor/non-rumor label at each time step/interval; 2) the stabilization component models the prediction stabilization process and determines when to stop at the earliest detection point. We will describe them with detail in this section.

\subsection{Rumor Detection Modeling}
A standard LSTM cell~\cite{hochreiter1997long} followed by a fully-connected layer is utilized for rumor detection in each interval. For any $(\mathbf{x}_i,t_i)\in X$, $\mathbf{x}_{i}$ can be turned into a vector $\mathbf{e}_{i}$ by a text representation method, e.g., TF-IDF~\cite{salton1988term}, CNN~\cite{kim-2014-convolutional}, BERT~\cite{devlin-etal-2019-bert}, etc.. Taking $\mathbf{e}_i$ as input, the LSTM cell gets the hidden state $\mathbf{h}_i=\text{LSTM}(\mathbf{e}_i)$ and forwards it through the fully-connected layer to perform prediction. The predicted class probability distribution of an instance at $t_i$ is calculated as $\mathbf{p}^r_i=\sigma(\mathbf{W}\mathbf{h}_i+\mathbf{b})$
and thus the predicted class is $\hat{y}_i=\text{argmax}(\mathbf{p}^r_i)$, where $\sigma(\cdot)$, $\mathbf{W}$ and $\mathbf{b}$ are sigmoid function, weight matrix and bias, respectively.

\subsection{Stabilization Process Modeling}
\textbf{Prediction Inverse (PI).}
Our rumor detection component keeps observing the posts stream and outputs a prediction sequence $\hat{y}_1,\hat{y}_2,\hat{y}_3,\ldots$ along the time steps. During the process, newly arrived posts may provide updated features rendering the next decision of rumor detection to invert from rumor to non-rumor or the other way round. Presumably, the predictions would get stabilized when sufficient clues are accumulated over time. By modeling such a process, we aim to fix the earliest time $t_i$ when the model can produce a stable prediction, meaning that there will be no expected inverses of prediction occurring from $t_i$ onward. Thus, we need to train the model for learning the expected future PIs to maximize its stability of prediction. 

As a future PI can occur unforseeably at any time, we introduce $I(t)$ as a cumulative count of PIs up to any time point $t$ in a \emph{continuous} time space to accommodate the uncertainty. Then a PI counts sequence $I(t_1),I(t_2),I(t_3)\ldots$ can be obtained from the prediction sequence. Clearly, we have $I(t_1)=0$, and $I(t_i)=I(t_{i-1})+1$ if $\hat{y}_i \ne \hat{y}_{i-1}$ or $I(t_i)=I(t_{i-1})$ otherwise. Additionally, we denote $\mathcal{H}_{t_{i}}=\{(I(t_1),t_1),\ldots,(I(t_{i-1}),t_{i-1})\}$ the history of PI counts up to $t\in (t_{i-1},t_{i}]$, and denote $\Delta I(t_{i},t_{j})=I(t_{j})-I(t_{i})$ the difference of PI counts between $t_i$ and $t_{j}$ for $j>i$.

\textbf{Neural Hawkes Process (NHP).} 
General Hawkes process~\cite{hawkes1971spectra} is a doubly stochastic point process for modeling sequence of discrete events in continuous time, which has been successfully applied in social media research, such as modeling the popularity of tweets~\cite{zhao2015seismic}, rumor stance classification~\cite{lukasik-etal-2016-hawkes}, fake retweeter detection~\cite{dutta2020hawkeseye}, and extracting temporal features for fake news detection~\cite{murayama2020fake}. Given a sequence of events $E_1,E_2,E_3,\ldots$ which occur at the corresponding time points $t_1,t_2,t_3,\ldots$ and $N(t)$ that is the number of events occurring up to time $t$, a uni-variate Hawkes process with a conditional intensity that indicates expected arrival rate of future events at $t$ is defined as~\cite{yang2013mixture}:
\begin{equation}
\lambda (t)=\mu +\int_{-\infty }^{t}\kappa (t-s)\mathrm{d}N(s)  
\label{pre:hp}
\end{equation}
where $\mu\ge0$ is the base intensity of event and $\kappa (\cdot)$ is a \emph{manually specified monotonic kernel function} that shows how the excitation from history decays with time. It assumes that arrived events can temporarily raise the probability of future events but the influence monotonically decays over time. However, this assumption is very strong which limits its ability for modeling complex dynamic point processes. In rumor diffusion, prediction inverse is the event influenced by many factors, such as what users express in historical and upcoming posts, which may bring tremendous uncertainty of prediction invalidating the monotonic decay assumption. Thus, we propose to adopt an NHP~\cite{mei-etal-2017-The} to capture the complex effects by utilizing a RNN with continuous-time LSTM (CTLSTM) to \emph{learn} the intensity function. CTLSTM extends the vanilla LSTM with an interpolation-like mechanism so that its hidden state for controlling intensity can be updated \emph{discontinuously} with each event occurrence and also evolves \emph{continuously} as time elapses towards the next upcoming event.

\textbf{Intensity Function Estimation.} 
We use NHP to model the dynamics of PI (i.e., event) and approximate a detection stability distribution over the  expected future predictions based on the sequence of historical and current predictions. As shown in Figure~\ref{fig:HEARD}, CTLSTM reads the current observation $(I(t_{i}),t_{i})$ to obtain $\lambda(t)$ for any $t>t_{i}$, so that the distribution over the expected PI counts can be approximated. The intensity function $\lambda(t)$ is controlled by a hidden state $\widetilde{\mathbf{h}}(t)$ as follow:
\begin{equation}\lambda(t)=f(\widetilde{\mathbf{W}}\cdot \widetilde{\mathbf{h}}(t))\label{pre:nhp}\end{equation} where $\widetilde{\mathbf{W}}$ is a weight matrix and $f(x) = \beta \log(1 + \exp(x/\beta))$ is the softplus function to obtain a positive intensity with a scale parameter $\beta$~\cite{mei-etal-2017-The}.

To model the unknown future for $t>t_{i}$ based on historical representation $\widetilde{\mathbf{h}}(t_{i})$, a new hidden cell vector $\widetilde{\mathbf{c}}(t)$ is introduced to control how $\widetilde{\mathbf{h}}(t)$ \emph{continuously} evolves over time. Specifically, $I(t_{i})$ is firstly transformed to a vector at $t_i$ by a fully-connected layer to join the updates in CTLSTM. Then, CTLSTM updates the hidden state $\widetilde{\mathbf{h}}(t)$ that has been evolving towards $\widetilde{\mathbf{h}}(t_{i})$ based on $\widetilde{\mathbf{c}}(t)$ for $t\in (t_{i-1},t_{i}]$. Thus, a richer representation of history can be learned by taking into account the dynamics of impact between the two consecutive observations. Meanwhile, to model the expected future PI count, $\widetilde{\mathbf{c}}(t)$ is updated to a new state with the current cell input, which is analogous to how vanilla LSTM updates hidden cell\footnote{The difference is that vanilla LSTM updates the hidden cell based on that of the previous time step while the update here is based on $\widetilde{\mathbf{c}}(t)$ for $t \to t_{i}$.}, and from the new state, $\widetilde{\mathbf{c}}(t)$ begins to continually approximate a target state that is defined by CTLSTM to represent an expected state of $\widetilde{\mathbf{c}}(t)$ for $t\to \infty$.

\textbf{Expected Stabilization.} $\lambda(t)$ indicates the expected instantaneous rate of future PIs from $t$ onward. We can predict the value of $I(t_{\infty})$ ($t_{\infty}$ denotes $t\to \infty$) and further determine an expected earliest stabilized observation $(I(t_{i^*}),t_{i^*})$, such that $t_{i^*}=\text{min}\{t_{i}\;|\; \Delta I(t_{i},t_{\infty})=0, i \in \{1,\ldots,|X|\}\}$. Hence, $t_{i^*}$ indicates the expected earliest time that the predictions remain unchanged after it. We approximate $\Delta I(t_{i},t_{\infty})$ by
\begin{equation}
  \Delta \hat{I}(t_{i},t_{\infty})=\int_{t_{i}}^{\infty} \lambda(t) \mathrm{d}t \label{N inf}  
\end{equation}
where $\hat{I}(\cdot)$ constantly denotes a predicted PI count.  We use Monte Carlo trick to handle all integral estimations~\cite{mei-etal-2017-The}. Given $t_{i^*}$, HEARD finally outputs $\hat{y}_{i^*}$ as the stable prediction for early rumor detection.

\subsection{HEARD Model Training}
We utilize a next observation $(I(t_{i+1}),t_{i+1})$ prediction task to train CTLSTM. With PIs as the indicator of model stability, CTLSTM aims to fit the sequence of observations $\mathcal{H}_{t_{i+1}}$ obtained from the predictions of the rumor detection module. 

Specifically, $\hat{I}(t_{i+1})$ can be obtained from $I(t_i)+\Delta\hat{I}(t_{i},t_{i+1})$, and the difference value $\Delta\hat{I}(t_{i},t_{i+1})$ is either 1 or 0 which can be predicted as $\Delta\hat{I}(t_{i},t_{i+1})=\text{argmax}(\mathbf{p}^c_i)$, where we infer $\mathbf{p}^c_i=\sigma(\mathbf{W'}\widetilde{\mathbf{h}}(t_i)+\mathbf{b'})$ with trainable parameters $\mathbf{W'}$ and $\mathbf{b'}$.
For predicting $t_{i+1}$, a density $p_{i}(t)$ is formulated as
\begin{equation}
p_{i}(t)=\lambda(t)\exp\left(-\int_{t_{i}}^{t}\lambda(s) \mathrm{d}s\right) 
\label{time_density}
\end{equation}
Then we use the minimum Bayes risk predictor for time prediction~\cite{mei-etal-2017-The}: 
\begin{equation}
\hat{t}_{i+1}=\mathbb{E}[t_{i+1}|t_i,\mathcal{H}_{t_i}] = \int_{t_i}^{\infty} p_i(t)t\mathrm{d}t \label{time_pred}
\end{equation}
where $\mathbb{E}[\cdot]$ is an estimator for choosing an optimal time point $t$ to minimize the expectation of risks.

The overall loss consists of three terms on rumor detection, expected earliest stable time and CTLSTM. Concretely, given an instance $C$ with input sequence $X= \{(\mathbf{x}_{i},t_{i})\}$, let $\mathbf{y}$ be the one-hot encoding of ground-truth label (i.e., rumor or not). At each time of observation $t_i$, the cross-entropy loss between prediction and ground truth is defined as $L^r_i=-\mathbf{y}\cdot \log\mathbf{p}^r_i$. For the expected earliest stable time, the loss at $t_i$ is defined as
\begin{equation}
L^e_i=|\Delta \hat{I}-\Delta I|-\log \left(1-\frac{i}{|X|}\right)
\label{loss stable}
\end{equation}
where the first term is the loss of $\Delta I(t_i,t_\infty)$\footnote{For simplicity, $\Delta I(t_i,t_{\infty})$ and $\Delta \hat{I}(t_i,t_{\infty})$ are denoted as $\Delta I$ and $\Delta\hat{I}$ in Eq.~\ref{loss stable}, respectively.} approximation in Eq.~\ref{N inf}, and the second term encourages the model to select an earliest time possible. The loss incurred from CTLSTM is given as
\begin{equation}
L^c_i=-\Delta \mathbf{I}\cdot \log\mathbf{p}^c_i+|\hat{t}_{i+1}-t_{i+1}|
\label{loss nhp}
\end{equation}
where $\Delta \mathbf{I}$ is the one-hot encoding of target $\Delta I(t_{i},t_{i+1})$, the first term is the loss of next prediction inverse and the second term is the loss of time prediction for next observation.

Our objective is to minimize the cumulative loss up to $t_{i^*}$ when the early detection decision is made: $L=\frac{1}{i^*}\sum_{i=1}^{i^*}L_i$, 
where $L_i=L^r_i+L^e_i+L^c_i$. We use stochastic gradient decent (SGD) mini-batch training over all training instances.

\section{Experiments and Results}
\subsection{Experimental Setup}
\textbf{Datasets.} 
We use BEARD, TWITTER~\cite{ma2016detecting} and PHEME~\cite{zubiaga2016learning} datasets in the evaluation. BEARD contains 1,198 rumors and non-rumors reported during 2015/03-2021/01 with around 3.3 million relevant posts. We hold out 20\% of instances for tuning, and the rest are randomly split with a ratio of 3:1 for training/test. Results are averaged over 5 splits. In Table \ref{tab:statistics}, we show the statistics of TWITTER, PHEME and BEARD datasets.

\begin{table}[t!]
\centering
\begin{adjustbox}{width={\linewidth},totalheight={!},keepaspectratio}%
\begin{tabular}{l|l|l|l|l}
\toprule[1.0pt]
\multicolumn{2}{l|}{Dataset} & Instances \# & Posts \#  & AvgLen (hrs)\\ 
\midrule[0.5pt]
\multirow{2}{*}{TWITTER}  & R  & 498          & 182,499   & 2,538\\
                          & N & 494          & 466,480  & 1,456 \\ 
\midrule[0.5pt]
\multirow{2}{*}{PHEME}    & R  & 1,972        & 31,230  & 10 \\
                          & N & 3,830        & 71,210   & 19\\
\midrule[0.5pt]
\multirow{2}{*}{BEARD}    & R  & 531          & 2,644,807 & 1,432\\
                          & N & 667          & 657,925   & 1,683\\ 
\bottomrule[1.0pt]
\end{tabular}
\end{adjustbox}
\caption{Statistics of datasets. R: Rumor; N: Non-rumor.}
\label{tab:statistics}
\end{table}

\textbf{Baselines.} We compare HEARD with four state-of-the-art baselines using their original source codes: 1) \textbf{BERT}~\cite{devlin-etal-2019-bert} is fine-tuned on the ``earliest rumor detection'' task~\cite{miao2021syntax}, in which the early detection strategy is to output a prediction using only the \emph{first} post of each instance. 2) \textbf{CED}~\cite{song2019ced} uses a fixed probability threshold to check if the prediction is credible for determining the early detection point. 3) \textbf{ERD}~\cite{zhou-etal-2019-early} uses a Deep Q-Network (DQN) to enforce the model to focus on early posts for determining the time point to stop and output the detection result. 4) \textbf{STN}~\cite{xia-etal-2020-state} use a time-evolving network to represent state-independent sub-events in posts sequence for classifying claims at each checkpoint. 

\subsection{Experimental Settings} \label{exp.s}
To balance the sequence length and granularity of time intervals, we pre-process posts sequences in the three datasets differently. We merge every 10 posts in BEARD and every 2 posts in TWITTER while we only merge the posts with the same timestamp in PHEME due to its generally short sequences. For each interval, both ERD and STN use pre-trained word embeddings to initialize the embedding matrix and fine-tune it in the training, while CED uses the TF-IDF method~\cite{salton1988term}, all of which follow the settings in the original papers~\cite{song2019ced,zhou-etal-2019-early,xia-etal-2020-state}. We also follow the original CED setting by using TF-IDF with 1,000 dimensions for representing the posts in each time interval. 

The hidden size of standard LSTM is set to 128 with the dropout rate of 0.1, and the size of CTLSTM is set to 64. We pad all the sequences in a batch to the same length as the longest one, with the batch size of 16. We use the Adam~\cite{kingma2014adam} with a learning rate of 2$e$-4 for optimization. To avoid overfitting, we add a L2 regularization with the weight of 1$e$-4. All values are fixed based on the validation set.

Our model HEARD is implemented using Pytorch\footnote{\url{https://pytorch.org/}}. We use the original source codes of all the baselines:  CED\footnote{\url{https://github.com/thunlp/CED}} and ERD\footnote{\url{https://github.com/DeepBrainAI/ERD}} are implemented with TensorFlow; BERT\footnote{\url{https://github.com/huggingface/transformers}} are implemented with Pytorch, and we use the base uncased pre-trained model; The code of STN is obtained directly from the authors of the original paper~\cite{xia-etal-2020-state} which is implemented with Pytorch. All the experiments are conducted on a server with 4*12GB NVIDIA GeForce RTX 2080 Ti GPUs.

\subsection{Evaluation Metrics}
We use the general classification evaluation metrics accuracy and F1-score together with several EARD-specific metrics (see below) for evaluation.

\textbf{Early Rate (ER)}~\cite{song2019ced} is defined as the utilization ratio of posts:
$ER=\frac{1}{|\mathbf{C}|} 
\sum_{C \in \mathbf{C}}\frac{i_{C}}{|C|}$
where $\mathbf{C}$ is the test set,  $i_{C}$ implies the early detection decision is made at the $i$-th post in instance $C$ and $|C|$ is the number of posts in it. \emph{Lower} ER means the model can detect rumors \emph{earlier}.

\textbf{Early Detection Accuracy Over Time (EDAOT)}. The metric of detection accuracy over time widely used~\cite{ma2016detecting,zhou-etal-2019-early,xia-etal-2020-state} is unsuitable for EARD models as it enforces a model to output a decision at each checkpoint whereas an EARD model can decide its own optimal decision point which may be earlier and more accurate than its output at the checkpoint. Our variant requires a model output result \emph{only} when it cannot make an early decision before a given checkpoint while both accuracy and average time of decisions will be presented. Specifically, given a set of checkpoints at time $\{t_1,...,t_k\}$, 
at the $j$-th checkpoint, the detection accuracy is 
$Acc_{t_{j}}=\sum_{C \in \mathbf{C}}\frac{\mathds{1}_{\hat{y}_{i^*}=y}}{|\mathbf{C}|}$
, where the binary function $\mathds{1}$ takes 1 if $\hat{y}_{i^*}=y$ or 0 otherwise. And the average time of decisions is 
$AvgT_{t_{j}}=\sum_{C \in \mathbf{C}}\frac{{t}_{i^*}}{|\mathbf{C}|}$, 
where $t_{i^*} \le t_{j}$, and $t_{i^*}=t_{j}$ if the model cannot make a decision before $t_{j}$.

\textbf{Stabilized Early Accuracy (SEA)} is a newly defined comprehensive metric considering accuracy, earliness and stabilization:  
\begin{equation}
\nonumber
\begin{split}
SEA=&~\frac{1}{3|\mathbf{C}|} 
\sum_{C \in \mathbf{C}}\left[\mathds{1}_{\hat{y}_{i^*}=y}+\left(1-\frac{i_{C}}{|C|}\right)\right.\\
&\left.+~\left(1-\frac{\Delta I(t_{i^*},t_{|X|})}{|X|-i^*}\right)\right]
\end{split}
\end{equation}
where 
the first term is the ratio of correctly predicted instances at the predicted time point $t_{i^*}$ indicating accuracy, the second term is the ratio of posts after $t_{i^*}$ indicating earliness, and the third term is the ratio of unchanged predictions after $t_{i^*}$ indicating stability. The value of SEA is bounded in $[0,1]$ and higher $\text{SEA}$ means better performance.

\subsection{Results and Analysis} \label{Experiments Results}
\begin{figure*}[t!]
    
    \subfigure[TWITTER]{
        \includegraphics[width=0.32\linewidth]{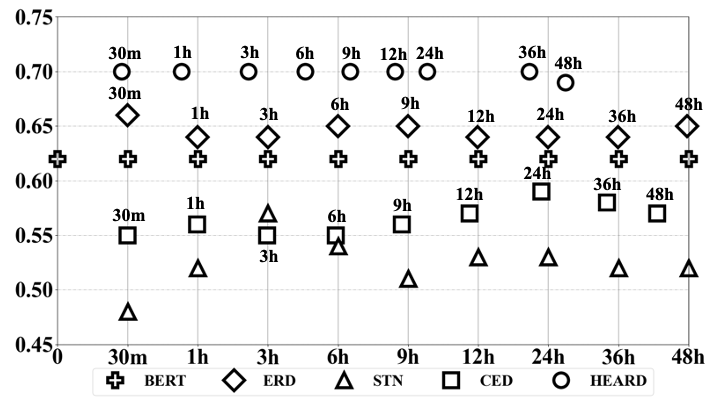}
        \label{Twitter}
    }
    \subfigure[PHEME]{
	    \includegraphics[width=0.32\linewidth]{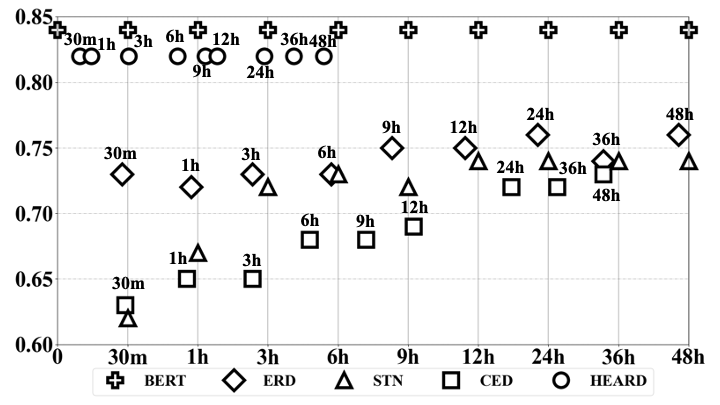}
        \label{PHEME}
    }
    \subfigure[BEARD]{
	    \includegraphics[width=0.32\linewidth]{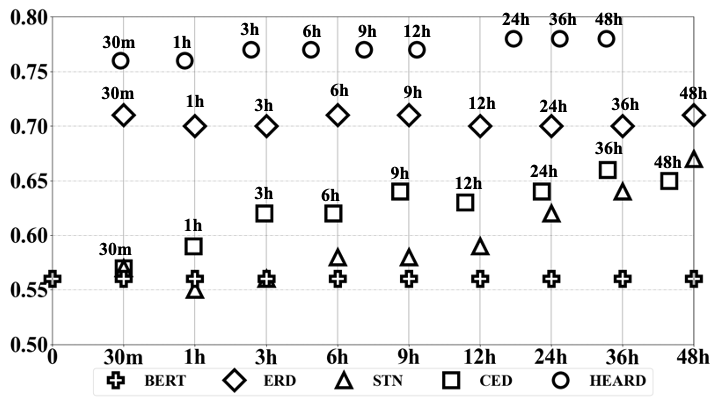}
        \label{BEARD}
    }
    \caption{Early detection accuracy over time within 48 hours. X-axis: timeline; Y-axis: accuracy; Vertical lines: checkpoints; Dots with shapes: decision points; Time above a dot: Deadline (i.e., checkpoint) set for the model.}
    \label{fig:acc over time}
\end{figure*}

The main results are provided in Table~\ref{tab:experiments}. We show the EDAOT results in Figure~\ref{fig:acc over time}.

\begin{table}[t!]
\small
\begin{adjustbox}{width={\linewidth},keepaspectratio}%
\begin{tabular}{c|c|c|c|c|c}
\toprule[1.0pt]
Dataset & Model     & Acc       & F1        & ER        & SEA  \\ 
\midrule[0.5pt]
\multirow{5}{*}{TWITT}
                    & BERT   & 0.623    & 0.599    & \textbf{0.026}     & 0.768     \\
                    & ERD   & 0.696   & 0.699    & 0.999     & 0.566    \\
                    & STN   & 0.682     & 0.649     & 1.000     & 0.561  \\
                    & CED   & 0.685     & 0.682     & 0.811     & 0.620       \\
                    & HEARD  & \textbf{0.716}& \textbf{0.714} & 0.348 & \textbf{0.789}    \\ 
\midrule[0.5pt]
\multirow{5}{*}{PHEME} 
                   & BERT   &\textbf{0.839}     & \textbf{0.820}    & \textbf{0.163}     &  0.830      \\
                    & ERD   & 0.784     & 0.753     & 0.976   & 0.602   \\
                    & STN   & 0.810     & 0.787     & 1.000     &  0.603   \\
                    & CED   & 0.800      & 0.695     & 0.884     & 0.638     \\
                    & HEARD  & 0.823 & 0.805 & 0.284 & \textbf{0.841}\\
\midrule[0.5pt]
\multirow{5}{*}{BEARD} 
                    & BERT   & 0.565     & 0.452     & \textbf{0.091}     & 0.758      \\
                    & ERD   & 0.709    & 0.708    & 1.000    & 0.570   \\
                    & STN   & 0.711     & 0.690     & 1.000     &  0.570    \\
                    & CED   & 0.769     & 0.740     & 0.674     & 0.689     \\
                    & HEARD  & \textbf{0.789} & \textbf{0.788} & 0.490 & \textbf{0.765}\\
\bottomrule[1.0pt]
\end{tabular}
\end{adjustbox}
\caption{Early rumor detection results.}
\label{tab:experiments}
\end{table}

\textbf{Results of Classification.}
        STN uses the entire timeline of each instance since it cannot automatically determine an early detection point. As shown in Table~\ref{tab:experiments}, however, it only achieves comparable Accuracy and F1 as ERD and CED and is much worse than HEARD, even though it was reported much better than CED on TWITTER in previous work~\cite{xia-etal-2020-state}\footnote{The CED performance reported in~\cite{xia-etal-2020-state} is an excerpt from~\cite{song2019ced} based on half of the TWITTER data, while they experimented STN using the full data.}. This implies that the early detection models are promising which can use a prior fraction of posts to achieve similar or much better results. It also suggests that capturing early-stage features is important to more accurate rumor detection. Only using the source post for detection, BERT gives worst Accuracy and F1 on TWITTER and BEARD, but it gets unexpectedly high performance on PHEME. We inspect this issue by following the prior analysis~\cite{schuster-etal-2019-towards} based on Local Mutual Information (LMI)~\cite{evert2005statistics} and Pointwise Mutual Information (PMI)~\cite{church-hanks-1990-word}. We find that the source posts in PHEME have spuriously much stronger correlation with the class labels. Appendix~\ref{bias} discusses such bias and the possible cause. This observation suggests data sampling bias exists in the existing dataset, and thus the model's decision based on the source post only can be misleading and insufficient.

\textbf{Results of ER and SEA.}
        Table~\ref{tab:experiments} also shows that HEARD consistently outperforms ERD and CED in large margin based on ER and SEA, indicating HEARD is more effective and stable. HEARD considerably improves CED by $57\%$, $68\%$ and $27\%$ in ER and by $27\%$, $32\%$ and $11\%$ in SEA on TWITTER, PHEME and BEARD, respectively. ERD's high ER scores entails that it can hardly make early decision, as this DQN-based model seems weak on its reward function, which gives a small penalty to continuation but a large one to termination with wrong predictions~\cite{zhou-etal-2019-early}, discouraging it from stopping early. BERT only uses the first post for detection which thus obtains the lowest ER. Note that a model being \emph{expectantly} stable at the time of decision means the prediction will remain unchanged even though new data could be seen by the model after that. To probe its stability, we enforce model to continue outputting predictions at the checkpoints later than its decision point. We can see that HERAD still outperforms BERT on SEA on all the datasets indicating it is more stable. 

\begin{figure*}[t!]
    \centering
    \subfigure[Germanwings]{
        \includegraphics[width=0.45\linewidth]{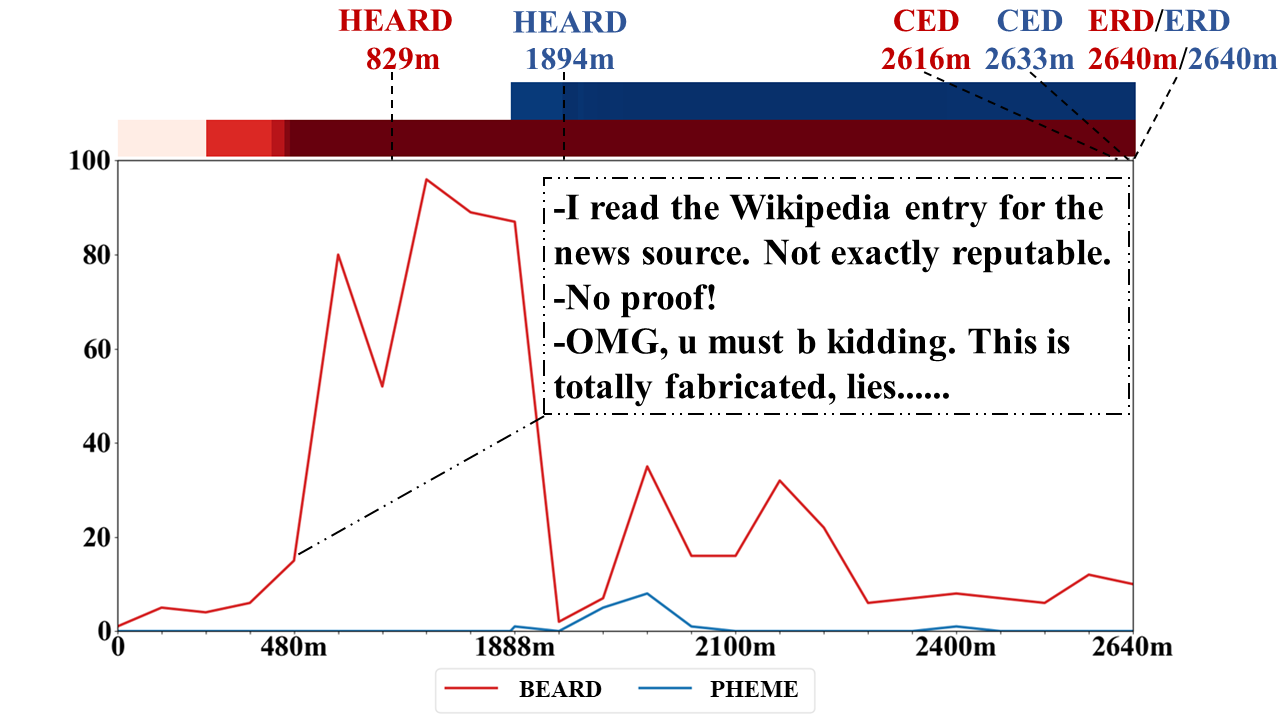}
        \label{PHEME3}
    }
    \subfigure[Darkest baby]{
	    \includegraphics[width=0.45\linewidth]{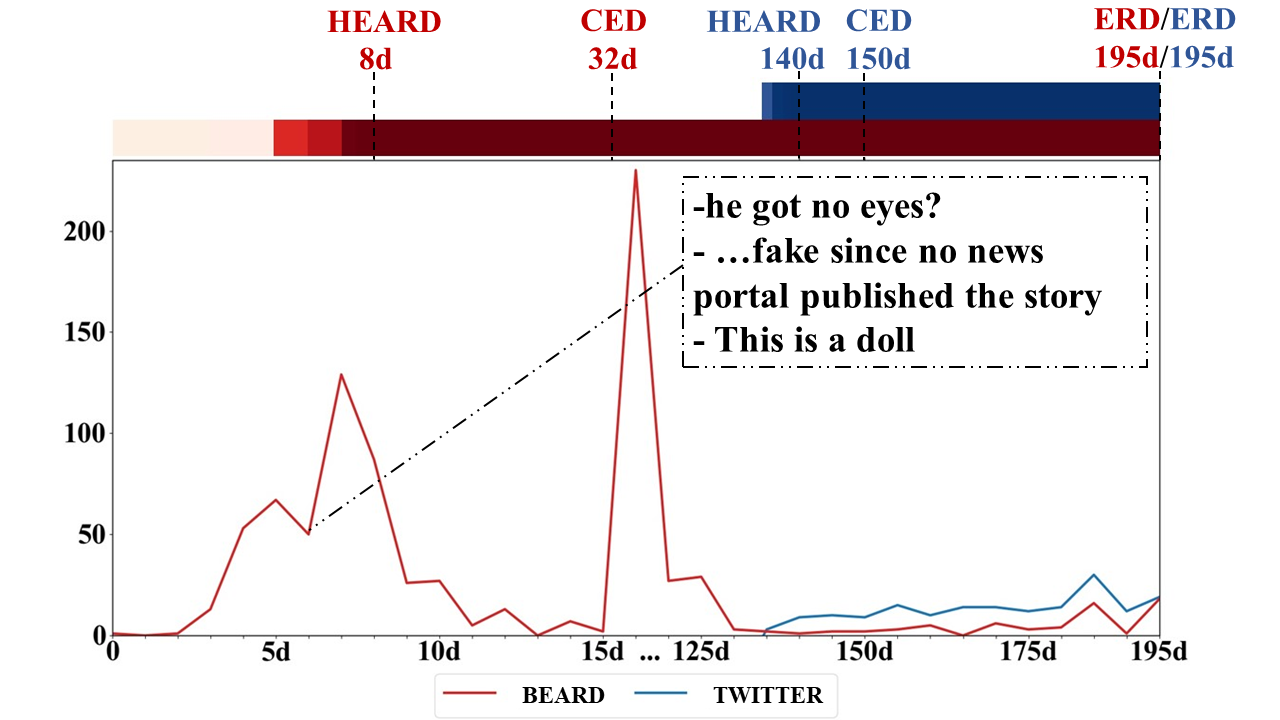}
        \label{TWITTER3}
    }
    \caption{(a) A rumor in PHEME on ``The co-pilot of Germanwings flight 9525 was Muslim convert''; (b) A rumor in TWITTER on ``A photograph shows the darkest baby in the world''. Y-axis: the number of collected posts; X-axis: timeline in minutes (m)/days (d); Colored bar: the semantic similarity between the posts in any time step and the posts in the decision time step, where the red bar denotes recollected data and the blue one denotes original data. The similarity ranges from 0.7 to 1.0, and the darker the color the more similar the posts are.}
    \label{fig:case}
\end{figure*}

\textbf{Results of EDAOT.}
        Figure~\ref{fig:acc over time} shows that HEARD is clearly superior over the baselines in accuracy, earliness and stability at the checkpoints. 
        ERD looks relatively stable but it hardly makes early detection due to aforementioned reason. Our conjecture is that its DQN module forces it to overly focus on the first few intervals while deferring the decision to the end due to the weak reward design, resulting in nearly no improvement even with more data. Note that in EDAOT evaluation a model should stop once it outputs a prediction for a given checkpoint (i.e., deadline). Thus, BERT reports all the predictions using only the first post, rendering the same accuracy at checkpoints. 
        Interestingly, HEARD can make especially fast decisions on PHEME which only uses a little less than 6 hours given a 48-hour deadline. The reason might be the average length of instances in PHEME is only around 14 hours as compared to over 1,000 hours in other two datasets, which HEARD can especially benefit from in the training. However, considering the fact that the PHEME (and TWITTER) dataset may fail to cover the real early information, such a very quick detection on it could be an illusion since the patterns the model actually uses for the decision are from the midst of rumor propagation.

\section{Case Study}

In this section, we give an intuitive illustration to reveal
1) why BEARD is more suitable for EARD task; and
2) how HEARD is more advantageous than the state-of-the-arts with automatic early time point determination.
Specifically, we recollect the relevant posts of rumor cases in PHEME and TWITTER using our data construction method for BEARD. To analyze the first question, we inspect the posts before and after the recollection as shown in Figure~\ref{fig:case}. In the original posts, the actual early-stage timelines of both cases are largely missing. Our recollected data can cover the important denial and questioning posts signaling rumors in the early stage. This observation indicates that our data collection method has improved coverage by recalling these actual early information which is not available in the existing datasets.

To analyze the second question, we display the detection outputs of different models before and after the recollection. HEARD consistently detects rumors with less time than CED and ERD since it automatically terminates when stabilization process estimates the expected future prediction inverse will not change. The correctness of such expectation could be reflected by the value of semantic similarity, which almost has no change with new posts coming in after the detection point, implying the future prediction results are expected stable.

\section{Conclusion and Limitation}
We introduce BEARD, a new benchmark dataset collected for early rumor detection, and propose a model called HEARD to perform stable early rumor detection based on neural Hawkes process. 
Experiments show that HEARD achieves overall better performance than state-of-the-art baselines.
Analysis entails BEARD is more suitable for the early rumor detection task. 

There are some cases that a rumor seems to be stable over an extended time period but is eventually refuted by an authoritative source later. It is because sensational discussions are more attractive to some social media users than facts, which leads to signals in support of the rumor are usually much stronger in social media, and the final correction often has only small engagement. In such cases, our model HEARD based on social media alone could be misled by the signals of only one source. In future, we plan to study incorporating authoritative sources of information to alleviate this phenomenon on social media, such as statements of official platforms, scientific sources, etc.. 

\section*{Acknowledgement}
This research is supported by the National Research Foundation, Singapore under its Strategic Capabilities Research Centres Funding Initiative and the Singapore Ministry of Education (MOE) Academic Research Fund (AcRF) Tier 1 grant. Any opinions, findings and conclusions or recommendations expressed in this material are those of the author(s) and do not reflect the views of funding agencies.

\bibliography{anthology,new}
\bibliographystyle{acl_natbib}

\appendix

\section{Corpus Construction Protocols} \label{ap:beard}

\subsection{Claims Collection} \label{Claims Information Retrieval}
Snopes\footnote{\url{https://www.snopes.com/}} is a well-known fact-checking website where fact-checkers manually collect check-worthy claims from multiple sources (e.g., social media, e-mail, news, etc.) and review each claim to report a decision in terms of ``Fact Checks'' (i.e., rumors being fact-checked) or ``News'' (i.e., non-rumors of no need to check). Further, the checkers verify the rumors and compose detailed fact-checking articles for justifying the veracity of rumors. 

We utilize Snopes to collect the claims for constructing our data instances in terms of rumors and non-rumors. There are around 10k+ claims on Snopes in total (up to the end of 2021), but majority of them have very limited exposure on Twitter. Specifically, we include the claims into our collection based on the following principles: (1) We only include the claims that were published in recent 6 years since they have relatively more complete exposure on Twitter (e.g., aged posts might be more likely to get deleted); (2) We only include the claims that have relevant posts on Twitter before the claim was officially debunked. As a result, we collect 531 rumor and 667 non-rumor claims reported during 2015/03-2021/01. For each claim, we scrape the text of article title, the claim, the debunking time and the veracity label. 

We also try to exclude posts that may leak the ground truth since a claim might have been fact-checked by other fact-checking websites, thus the truth might have been referenced in the social media posts, such as the rumor example in TWITTER shown in Table~\ref{tab:intro}. To minimize the chance of ground truth leakage, therefore, we match each claim from Snopes across the claims from a handful set of fact-checking websites including FactCheck.org\footnote{\url{https://www.factcheck.org/}} and PolitiFact.com\footnote{\url{https://www.politifact.com/}} to get the possible earliest official debunking time, from which we begin collecting the relevant posts backwards in time. 

\subsection{Query Construction} \label{Query Construction}
For each claim, we construct a set of high-quality search queries for Twitter search to diversely cover relevant posts. Firstly, we concatenate the title and claim of an article followed by stop words removal to filter out noise, resulting in an original query. Since the original query might be long and harm the diversity of search results, we carry out the following ad hoc operations to shorten the query for maintaining maximum useful information and possibly broadening the coverage of retrieved posts: (1) We perform synonym replacement with Natural Language Toolkit (NLTK) \cite{bird2009natural} to create a variant of the query for each replacement; (2) For each variant obtained in (1), we use Google Search API\footnote{\url{https://developers.google.com/custom-search/v1/overview}} to search for this altered query and rank by the frequencies of the \textit{highlighted words} that are hit in the top-100 searched snippets; (3) We then shorten the original query and its variants obtained in (2) by removing the highlighted words that are contained in the queries one by one starting from the low-frequency words, while keeping the remaining words after each removal as a variant of the query, until the shortest variant is left with three words. Note that here we perform named entity recognition by NLTK on the original query and the words that are parts of named entities will not be removed as they are useful for search.

\subsection{Iterative Twitter Search} \label{Iterative Twitter search}
To prevent early quit of iteration caused by the intermittent sparse distribution of posts along the timeline, we manually adjust the values of $M$ and $N$ by trial and error with different instances to gather as much early posts as we can. We also manually check the search results based on a sample of instances using different settings of the termination threshold, i.e., the ratio of the number of gathered posts in each iteration over that in the previous iteration. We observe that too high threshold hinders early posts to be searched out, but too low threshold tends to introduce more noise. We finally set 1\% as the termination threshold by trading off earliness and noise.

\subsection{Posts Collection} \label{Posts Collection}
Rather than merging the conversations led by different root posts regarding the same claim into a large conversation, it is more realistic to remain them naturally separated since the conversations originate from different sources and the merge may introduce unnecessary bias. And we drop all the posts that are published after the official debunking time of each claim to avoid ground truth leakage.

\section{Experimental Details}

\begin{table}[t]
\centering
\begin{adjustbox}{width={0.8\linewidth},keepaspectratio}%
\begin{tabular}{l|l|l|l}
\toprule[1.0pt]
Dataset & Word & LMI$\cdot10^{-6}$     & PMI        \\ 
\midrule[0.5pt]
\multirow{5}{*}{PHEME}
                    & breaking   & 2,720   & 0.75 \\
                    & hostages   & 2,569     & 0.66     \\
                    & soldier  & 2,378     & 1.00    \\
                    & shot   & 2,308     & 0.79    \\
                    & cafe  & 2,283 & 0.74 \\ 
\midrule[0.5pt]
\multirow{5}{*}{TWITT}
                    & not   & 1,214   & 0.38 \\
                    & black   & 780     & 0.45     \\
                    & flag   & 745     & 0.49    \\
                    & trump   & 637     & 0.39    \\
                    & baby  & 589 & 0.53 \\ 
\midrule[0.5pt]
\multirow{5}{*}{BEARD}
                    & he   & 1,246   & 0.38 \\
                    & they   & 722     & 0.26     \\
                    & his   & 665     & 0.26    \\
                    & by  & 598 & 0.19\\  
                    & biden   & 580     & 0.51    \\
\bottomrule[1.0pt]
\end{tabular}
\end{adjustbox}
\caption{Top-5 LMI-ranked words of source posts correlated with class label in three datasets.} 
\label{tab:LMI}
\end{table}

\subsection{Bias Evaluation on Datasets} \label{bias}

As mentioned in Section~\ref{Experiments Results}, we utilize LMI and PMI to examine the data sampling bias in the existing datasets and BEARD. The top-5 LMI-ranked words of PHEME, TWITTER and BEARD are shown in Table~\ref{tab:LMI}. The top-5 LMI-ranked words of PHEME have much higher LMI and PMI than the other two datasets indicating the high correlations between the words of source posts and the label. Meanwhile, the words in PHEME, e.g., `breaking', `hostages', `shot', etc., are more eyes-catching comparing to the top ranked words in other two datasets. These idiosyncrasies might be introduced by the construction method of PHEME in a sense that journalists might see a timeline of posts about the breaking news and then annotate source posts of conversations, which are easily utilized by BERT to obtain high classification performance. Some salient bias also exists in TWITTER dataset evidenced as some top words with high LMI and PMI, such as `black (lives)', `flag' and `trump', while we do not observe such bias among the top words in BEARD.

\end{document}